\documentclass[preprint,12pt,a4paper]{elsarticle}
\usepackage{amsmath}
\usepackage{amssymb}
\usepackage{bm}
\usepackage{array}
\usepackage{booktabs}
\usepackage{enumerate}
\usepackage{graphicx}
\usepackage{subcaption}
\usepackage{float}
\usepackage{comment}
\usepackage{hyperref}

\biboptions{numbers,sort&compress}

\hypersetup{
  colorlinks=true,
  linkcolor=blue,
  citecolor=blue,
  urlcolor=blue,
  pdftitle={A Transferable Autologistic Model for Predicting Rare Failures in Heterogeneous Equipment},
  pdfauthor={Islam Benamirouche, Djemel Ziou, Feriel Fass}
}

\newcolumntype{L}[1]{>{\raggedright\arraybackslash}p{#1}}

\begin{document}

\let\WriteBookmarks\relax
\renewcommand{\floatpagefraction}{.8}
\renewcommand{\textfraction}{.05}

\journal{Reliability Engineering \& System Safety}

\begin{frontmatter}

\title{A Transferable Autologistic Model for Predicting Rare Failures in Heterogeneous Equipment}

\author{Islam Benamirouche}
\ead{Islam.Benamirouche@usherbrooke.ca}

\author{Djemel Ziou\corref{cor1}}
\ead{Djemel.Ziou@usherbrooke.ca}

\author{Feriel Fass}
\ead{Feriel.Fass@usherbrooke.ca}

\affiliation{
  organization={Département d'informatique, Université de Sherbrooke},
  addressline={2500 boulevard de l'Université},
  city={Sherbrooke},
  postcode={J1N 3C6},
  state={QC},
  country={Canada}
}

\cortext[cor1]{Corresponding author}

\begin{abstract}
Predicting failures before they occur remains a major challenge in predictive maintenance, particularly when failures are rare, when equipment of the same family differ in sensor configurations, and when the goal is anticipation rather than diagnosis of an already observed fault. This paper proposes a common-to-target probabilistic model that learns shared failure-related patterns across a family of heterogeneous equipment and adapts parsimoniously to target equipment. The model explicitly accounts for sensor heterogeneity, operating context, and degradation dynamics to produce calibrated failure-probability estimates suitable for maintenance planning. Its performance is evaluated on a synthetic refrigerator dataset comprising 27 simulated refrigerators with varying sensor configurations, operating conditions, and failure types, providing a controlled setting for assessing common-model generalization to target equipment.
\end{abstract}


\end{frontmatter}

\section{Introduction}
\label{sec:intro}

Many equipment is increasingly monitored through sensors that record time-dependent measurements, making it possible to move from reactive maintenance, where intervention occurs after failure, to predictive maintenance, where the objective is to anticipate failures before they interrupt normal operation. This is particularly important for equipment that operate repeatedly or continuously, because unexpected failures can cause service interruption, loss of stored goods, safety issues, or additional maintenance costs. In this paper, a failure is defined as an abnormal condition affecting one or more components, after which the equipment no longer behaves as expected and maintenance action may be required. Before failure occurrence, degradation may progressively affect equipment behavior and modify one or more sensor-output variables. Abnormal patterns may therefore appear in temperature, pressure, electrical current, vibration, or component-state measurements before the failure becomes observable, so failure prediction consists of estimating, at a given observation time, whether a failure will occur.

This prediction problem is difficult for two main reasons. First, failures are rare: only a small number of observations are associated with future failures, while most observations correspond to normal operation \cite{carvalho2019slr,zonta2020industry4review,syed2025timeseriesreview}. This creates strong class imbalance and makes it difficult to learn reliable pre-failure patterns. Second, degradation may be distributed across several time-dependent measurements rather than appearing in a single sensor-output variable \cite{nunes2023manufacturing,leukel2021failureprediction,dangut2022rareaircraft}. The problem becomes more complex when the model must be applied across heterogeneous equipment from the same family. An equipment family comprises related equipment that perform the same function and follow similar physical principles while possibly differing in sensor configuration, calibration, operating conditions, and usage. These differences make direct model application difficult, especially when the available sensor-output variables or their scales differ across equipment.

This paper presents a common-to-target probabilistic model for failure prediction across heterogeneous equipment. The model is learned from training equipment and then specialized to target equipment using limited target data. Here, target equipment denotes equipment from the same family that is excluded from common-model training and used afterward for adaptation and evaluation. The main contributions are a common-to-target model for heterogeneous equipment; a representation combining a common sensor space, availability masks, and context variables; a probabilistic autologistic model based on a supervised latent vector, class weighting, and maximum a posteriori (MAP) parameter estimation; a target-specific adaptation strategy that uses the common model as prior information; and an experimental evaluation based on maintenance-oriented metrics assessing detection, lead time, and false alerts.

The proposed approach is evaluated on a controlled synthetic dataset generated from a family of 27 refrigerators. Refrigerators are used as a case study because they operate continuously, generate multivariate sensor-output time series, and exhibit heterogeneous sensor configurations, variable operating conditions, and several degradation mechanisms. Seventeen refrigerators are used as training equipment to estimate the common model, while ten are used as target equipment for common-model prediction, target-specific adaptation, and evaluation. The experiments assess whether the common model can be applied to target equipment and whether target-specific adaptation improves failure detection and reduces false alerts. The remainder of the paper is organized as follows. Section~\ref{sec:related} reviews related work. Section~\ref{sec:problem} defines the failure prediction problem and the representation of heterogeneous equipment data. Section~\ref{sec:methodology} presents the prediction model. Section~\ref{sec:global_local} explains how the common model is specialized to target equipment. Section~\ref{sec:simulator} describes the synthetic dataset and the evaluation protocol. Section~\ref{sec:results} reports and discusses the results. Section~\ref{sec:conclusion} concludes the paper.

\section{Related work}
\label{sec:related}

Since this paper focuses on failure prediction, this section reviews prior work related to four aspects of the proposed problem, namely failure prediction with rare failures, transfer across related equipment, heterogeneous sensor configurations, and probabilistic temporal modeling. Several reviews show that predictive maintenance has been studied through anomaly detection, fault detection, diagnosis, remaining useful life estimation, and failure prediction \cite{carvalho2019slr,zonta2020industry4review,syed2025timeseriesreview}. However, failure prediction has received less attention than fault detection, diagnosis, and remaining useful life estimation, although it is directly relevant when maintenance decisions must be made before a failure occurs \cite{leukel2021failureprediction}. Recent reviews also emphasize the difficulty of applying data-driven predictive maintenance when failures are rare, data are heterogeneous, and operating conditions vary \cite{nunes2023manufacturing,ucar2024trustworthy,moleda2023review}. In this work, the difficulty is increased because the prediction model must be applied across heterogeneous equipment from the same equipment family.

A first difficulty is that failures are rare. Only a small proportion of observations are followed by a failure, which creates strong class imbalance and can make model estimation unstable. Weighting methods have therefore been studied to reduce the effect of imbalance in rare-failure settings \cite{hewieghting2021frontiers}. Failure prediction with rare failures has also been studied in household equipment using co-evolving time series, showing the relevance of probabilistic temporal models for this problem \cite{mecheri2024rareevents}. Other approaches estimate failure probability over a future period to support maintenance planning \cite{fu2023failureprobability,chen2023failuredecision}. Rare failures have also been considered in aircraft predictive maintenance, where the main difficulty is to identify failure-related patterns from a small number of failure observations \cite{dangut2022rareaircraft}. These works formulate maintenance as future failure-probability estimation rather than only as classification of the current state. However, many of them assume a fixed sensor configuration or a single operating setting. This limits their use when the same prediction model must be applied to related equipment with different sensor configurations, operating conditions, or usage profiles.

Household equipment, thermal equipment, and building facilities provide a relevant application context for this problem because they are monitored by sensor-output time series and may exhibit progressive degradation before failure. Existing work has proposed datasets and monitoring models for identifying maintenance needs in home appliances \cite{fonseca2023homeappliancedataset}. Predictive maintenance has also been studied in building facilities and fleet management, where operating context and usage conditions influence maintenance decisions \cite{bouabdallaoui2021buildingpdm,giannoulidis2023fleet,zeng2023fleetpdm}. Refrigeration systems, vaccine refrigerators, chillers, and HVAC systems have been studied mainly for fault detection and diagnosis \cite{kim2021chillerfdd,soltani2022refrigerationfdd,abhiraman2023vaccinerefrigeration,wang2026hvacdt}. IoT-based monitoring has also been used for refrigeration equipment failure detection or prediction \cite{kokosinski2025rst}. These works show that sensor-output time series can support maintenance decisions for household and thermal equipment. Nevertheless, most of them focus on detecting or diagnosing an existing fault, or on a specific equipment configuration. They do not generally address probabilistic prediction of future failures across a family of heterogeneous equipment.

This limitation motivates transfer across equipments of the same equipment family. Training an independent model for each target equipment is often unrealistic because failures are rare and target data may contain few failure observations. Transfer learning has therefore been introduced in predictive maintenance to reuse information learned from training equipment \cite{samanazari2023tlreview,yadav2024iiottl}. Partial transfer learning, few-shot learning, and self-supervised representation learning have also been studied to reduce the amount of target data required for fault diagnosis under limited data conditions \cite{li2020partialtl,wang2021fsm3,wang2024selfsupervisedfewshot,wang2024noisyfewshot}. Applications include chillers, building energy systems, wind turbines, and offshore wind turbines \cite{vandesand2021heterogeneouschillers,li2023bescross,li2021windtl,trizoglou2021xgboost}. Domain-adaptation methods have also been proposed for HVAC fault diagnosis under changing operating conditions \cite{ghalamsiah2025can,ghalamsiah2026graphuda}. Digital twins and physics-guided transfer have been used to reduce the gap between simulated and physical equipment \cite{wang2023digitaltwintl,lohalekar2025physicsguidedtl}. Most of these works, however, address fault diagnosis or fault-type classification rather than probabilistic failure prediction with limited target data.

A specific obstacle to transfer is that equipments of the same equipment family may not share the same sensor configuration. Some sensor outputs may be permanently absent by design, while others may be temporarily unavailable because of sensor failure, communication loss, or invalid readings. This issue is related to sensor-set mismatch, feature heterogeneity, and missing data in multivariate time series. Feature-alignment and domain-adaptation methods aim to make representations more comparable across training equipment and target equipment. Sensor ontologies, such as SSN, provide formal descriptions of sensors, observations, and measured quantities \cite{haller2019ssn}. Knowledge graph approaches can represent equipment structure and maintenance information in a consistent form \cite{zhao2025kgom}. Missing-data-aware models such as GRU-D show that missingness patterns can contain useful information and should not always be treated as noise \cite{che2018grud}. These works motivate explicit representations that distinguish measured sensor-output values from unavailable values, since an unavailable sensor output is not equivalent to a zero measurement and a permanently absent sensor is not the same as a temporary missing value.

Once the equipment data are represented in a common form, the prediction model must estimate failure probabilities rather than only class labels, because maintenance decisions depend on the predicted probability of a future failure. Data-driven predictive maintenance models include support vector machines, tree-based models, neural networks, and autoencoder-based models \cite{Datadriven2019IEEE}. Dimensionality-reduction methods can be used to reduce high-dimensional input spaces \cite{wani2025dimensionality}, while supervised autoencoders can learn latent representations that preserve information useful for prediction \cite{jarrett2020targetembedding}. Logistic regression and related binary probabilistic models provide interpretable failure probabilities. Bayesian and maximum a posteriori (MAP) formulations can incorporate prior information, which is useful when target data contain few failure observations. Temporal dependence can be introduced through autologistic structures, where previous predicted probabilities influence the current predicted failure probability \cite{betterautologistic,autologisticusecase1}. Related mixed time-series models also represent temporal dependence through low-dimensional probabilistic structures \cite{debaly2021mixedtimeseries}. These models provide statistical foundations for failure prediction with rare failures, but they do not by themselves address the joint problem of transfer, heterogeneous sensing, and target-specific adaptation.

Overall, the literature provides important points. These include failure prediction under class imbalance, data-driven monitoring of household and thermal equipment, transfer across related equipment, representation of heterogeneous sensor information, supervised latent representation learning, and probabilistic temporal prediction. These points are, however, generally studied separately. To the best of our knowledge, the joint problem of probabilistic failure prediction across heterogeneous equipment with rare failures and target-specific adaptation has not been extensively studied within a single common-to-target model. This paper addresses this gap by learning a common model from training equipment and specializing the final prediction model to each target equipment using limited target data.

\section{Methodology and representation of heterogeneous equipment data}
\label{sec:problem}

We consider a family of equipment monitored by sensors at successive observation times. For each equipment $e$, the index $t = 1, \ldots, N_e$ denotes the observation time. Indeed, each equipment is fitted with $m_e$ sensors that produce time-dependent measurements of physical and operational quantities, such as temperature, pressure, humidity, electrical current, vibration, and component state. Each sensor $j$ produces a quantity $X_{t,j}^{(e)}$ at observation time $t$, which may be real-valued or binary, provided that the corresponding measurement is available and valid. A sensor may be absent from the equipment by design, or it may become temporarily unavailable over a time interval $[t_1,t_2]$ due to a sensor failure, communication failure, or invalid reading. In both cases, no valid measurement is available and the corresponding output is set to zero.

Let $X_t^{(e)} = \bigl(X_{t,1}^{(e)}, \ldots, X_{t,m_e}^{(e)}\bigr)^\top \in \mathbb{R}^{m_e}$ denote the sensor-output vector at time $t$. This vector contains all sensor-related and derived variables that may differ from one equipment to another, since different equipment are not necessarily fitted with the same set of sensors.

In addition to the sensor outputs, each equipment operates within a context that gathers external operating conditions and historical maintenance information. These variables may be of different statistical types: real-valued, binary, or count. Let $C_t^{(e)} = \bigl(C_{t,1}^{(e)}, \ldots, C_{t,d_c}^{(e)}\bigr)^\top \in \mathbb{R}^{d_c}$ denote the context vector at time $t$. This vector contains information that cannot be captured by sensors, such as the time since last maintenance, the equipment's age, and the cumulative number of failures recorded before time $t$. These variables are defined for every equipment in the family.

Equipment failure incurs high operational costs, including repair expenses and revenue losses during the downtime period. Anticipating failures enables scheduled interventions and reduces these costs. We therefore seek to estimate, at each time $t$, whether at least one failure will occur within the future horizon $]t, t+H]$, where $H > 0$ is the prediction horizon. The value of $H$ influences the maintenance strategy. A small $H$ may lead to unstable or erroneous predictions, whereas a large $H$ may reduce operational feasibility. To formalize this prediction task for a prescribed horizon $H$, we introduce a binary variable that flags whether at least one failure materializes within the future window. Let $Y_t^{(e)}$ denote the binary random variable associated with the occurrence of at least one failure of equipment $e$ in the future horizon $]t,t+H]$. 

\begin{equation} 
  Y_t^{(e)} = \begin{cases} 1, & \text{if at least one failure occurs in } ]t, t+H], \\ 0, & \text{otherwise.} \end{cases} 
  \label{eq:future_target} 
\end{equation}

The prediction task consists in using the sensor outputs and the context available at time $t$ to estimate the probability of future failure within $]t, t+H]$: 

\begin{equation} 
  p_t^{(e)} = \mathbb{P} \left( Y_t^{(e)} = 1 \mid X_t^{(e)}, C_t^{(e)} \right). 
  \label{eq:prediction_target} 
\end{equation}

Before specifying the form of this probability, we must address the structural differences between equipment in the same family. In this work, a family of equipment refers to a set of equipment that serve the same general purpose while possibly differing in size, capacity, technical properties, sensor availability, and usage conditions. These differences manifest in three ways. First, equipments of the same family may not share the same sensor configuration. For example, in the case of fridges, one fridge may be fitted with vibration, humidity or pressure sensors, whilst another may only have thermal and electrical sensors. Second, even when two sensors measure the same physical quantity, the recorded variables may differ in terms of name, unit, or scale. For instance, compressor activity may be represented as a binary on/off signal on one equipment and as an electrical-current measurement on another. Third, equipment may operate under different conditions, such as indoor or outdoor installation, or frequent versus infrequent door openings, which affect the degradation dynamics.

For this reason, we need to build a common feature space for the  sensor-output vector $X_t^{(e)}$ for all equipments of same family and a common space for the context $C_t^{(e)}$ for all equipements of this family, that there are $d$ different sensors embedded in the equipements of the family. At each time $t$, the The sensor-output vector of equipment $e$ is rewritten in the common feature space as $\widetilde{X}_t^{(e)} = \bigl(\widetilde{X}_{1,t}^{(e)}, \ldots, \widetilde{X}_{d,t}^{(e)}\bigr)^\top \in \mathbb{R}^d$, with \begin{equation} \widetilde{X}_{i,t}^{(e)} = \begin{cases} \text{the valid measurement at time } t, & \text{if } M_{i,t}^{(e)} = 1, \\ 0, & \text{if } M_{i,t}^{(e)} = 0, \end{cases} \label{eq:common_sensor_def} \end{equation} where $M_{i,t}^{(e)} \in \{0,1\}$ indicates whether a valid measurement is available. For each equipment $e$ and each time $t$, the availability mask is $M_t^{(e)} = \bigl(M_{1,t}^{(e)}, \ldots, M_{d,t}^{(e)}\bigr)^\top \in \{0,1\}^d$. The pair $\bigl(\widetilde{X}_{i,t}^{(e)},\, M_{i,t}^{(e)}\bigr)$ resolves the ambiguity inherent in a zero-filled value. An undefined value with $M_{i,t}^{(e)} = 0$ may indicate sensor unavailability (structural absence if permanent, transient if temporary), whereas a zero with $M_{i,t}^{(e)} = 1$ is a valid sensor measurement. Using these vectors together with the context vector $C_t^{(e)}$, the complete input vector of dimension $d_\Phi = 2d + d_c$ for equipment $e$ at time $t$ is $\Phi_t^{(e)} = \bigl(\widetilde{X}_t^{(e)\top}, M_t^{(e)\top}, C_t^{(e)\top}\bigr)^\top$. This vector describes the state of equipment $e$ at a single observation time $t$. However, a future failure is not generally indicated by a single instantaneous measurement. It may instead be indicated by the recent evolution of the sensor-output time series and the context variables. For example, a temperature value may remain within its nominal range at time $t$, while its recent upward trend, combined with intense compressor activity, may signal a developing failure. We formulate the hypothesis that the prediction depends on $q$ recent input vectors. Let $\Phi_{t,q}^{(e)} = \bigl(\Phi_{t-q}^{(e)\top}, \ldots, \Phi_{t-1}^{(e)\top}, \Phi_t^{(e)\top}\bigr)^\top \in \mathbb{R}^{(q+1)d_\Phi}$ denote the concatenated input vector. When $q$ or $d_\Phi$ is large, estimating the failure probability directly from $\Phi_{t,q}^{(e)}$ leads to a high-dimensional estimation problem, often referred to as the curse of dimensionality. Since failure examples are rare, we map $\Phi_{t,q}^{(e)}$ to a lower-dimensional latent vector.

Linear dimensionality-reduction methods, such as principal component analysis (PCA), could be used for this purpose \cite{wani2025dimensionality}. However, the objective here is not only to compress the input window. The latent vector must also preserve information that is discriminative for the future-failure label $Y_t^{(e)}$, while reducing the effect of non-informative variables and lowering algorithmic complexity. This is achieved through a supervised autoencoder, which learns a compressed representation by jointly minimizing a reconstruction loss and a supervised prediction loss \cite{jarrett2020targetembedding}.

The supervised autoencoder consists of an encoder, a decoder, and a supervised prediction head. The encoder $f_{\mathrm{enc}}$ maps the concatenated vector $\Phi_{t,q}^{(e)}$ to a lower-dimensional latent vector: 

\begin{equation}
\begin{aligned}
Z_t^{(e)} &= f_{\mathrm{enc}}\!\left(\Phi_{t,q}^{(e)}\right), \\
f_{\mathrm{enc}} &: \mathbb{R}^{(q+1)d_\Phi} \rightarrow \mathbb{R}^{d_z}, \\
d_z &\ll (q+1)d_\Phi.
\end{aligned}
\label{eq:encoder}
\end{equation}

The decoder $f_{\mathrm{dec}}$ reconstructs the concatenated vector from the latent vector: \begin{equation} \widehat{\Phi}_{t,q}^{(e)} = f_{\mathrm{dec}}\!\left(Z_t^{(e)}\right), \qquad f_{\mathrm{dec}} : \mathbb{R}^{d_z} \to \mathbb{R}^{(q+1)d_\Phi}. \label{eq:decoder} \end{equation} The reconstruction loss measures how much information from the original input window is preserved in the latent vector. It is defined as the mean squared reconstruction error: \begin{equation} \mathcal{L}_{\mathrm{rec}}^{(e,t)} = \frac{1}{(q+1)d_\Phi} \| \Phi_{t,q}^{(e)} - \widehat{\Phi}_{t,q}^{(e)} \|_{2}^{2}. \label{eq:reconstruction_loss} \end{equation} The normalization by $(q+1)d_\Phi$ makes this loss a mean error per input dimension.

The second component is an auxiliary logistic prediction head. It is applied to the latent vector $Z_t^{(e)}$ during supervised autoencoder training in order to evaluate whether this latent vector contains information predictive of the future-failure label $Y_t^{(e)}$. This head produces an auxiliary probability:

\begin{equation}
\widehat{p}_{\mathrm{aux},t}^{(e)} = \sigma\!\left(\alpha_0 + \alpha^\top Z_t^{(e)}\right), \qquad \sigma(u) = \frac{1}{1+\exp(-u)}.
\label{eq:auxiliary_logistic_head}
\end{equation}

Here, $\alpha_0$ is the intercept and $\alpha$ is the coefficient vector of the auxiliary logistic head. These parameters are estimated jointly with the encoder parameters by minimizing the negative log-likelihood of the observed labels:

\begin{equation}
\mathcal{L}_{\mathrm{sup}}^{(e,t)} = -Y_t^{(e)} \log\!\left(\widehat{p}_{\mathrm{aux},t}^{(e)}\right) - \left(1-Y_t^{(e)}\right) \log\!\left(1-\widehat{p}_{\mathrm{aux},t}^{(e)}\right),
\label{eq:supervised_loss}
\end{equation}

For equipment $e$, the supervised autoencoder is trained over the $N_e$ observation times by minimizing the following loss function:
\begin{equation}
\mathcal{L}_{\mathrm{SAE}} = \frac{1}{\sum_{e} N_e} \sum_{e} \sum_{t=1}^{N_e} \left[ \lambda_{\mathrm{rec}} \mathcal{L}_{\mathrm{rec}}^{(e,t)} + \lambda_{\mathrm{sup}} \mathcal{L}_{\mathrm{sup}}^{(e,t)} \right],
\label{eq:sae_loss}
\end{equation}
where $\lambda_{\mathrm{rec}}\geq 0$ and $\lambda_{\mathrm{sup}}\geq 0$ are hyperparameters controlling the relative importance of reconstruction and supervised prediction. The reconstruction term preserves the structure of the concatenated vector $\Phi_{t,q}^{(e)}$, while the supervised term ensures that the latent vector $Z_t^{(e)}$  remains discriminative for failure prediction.

After training, the latent vector $Z_t^{(e)}$ is used as the input to the probabilistic prediction model, so that the prediction problem becomes
\begin{equation}
p_t^{(e)} = \mathbb{P}\!\left(Y_t^{(e)}=1 \mid Z_t^{(e)}\right).
\label{eq:risk_latent}
\end{equation}
The following section defines the parametric form of this probability.

\section{Failure prediction and parameter estimation}
\label{sec:methodology}

The previous section defined the latent vector $Z_t^{(e)} \in \mathbb{R}^{d_z}$, which summarizes the memory of the random process $(\Phi_t^{(e)}, \Phi_{t-1}^{(e)}, \ldots, \Phi_{t-q}^{(e)})$. The present section defines the parametric model of the failure probability.

As defined in Eq.~\eqref{eq:future_target}, $Y_t^{(e)}$ denotes the binary random variable associated with the occurrence of at least one failure in the future horizon $]t,t+H]$. Since $Y_t^{(e)}$ is binary, its conditional distribution given $Z_t^{(e)}$ is modeled as a Bernoulli distribution:

\begin{equation}
\begin{gathered}
\mathbb{P}\!\left(Y_t^{(e)} = 1 \mid Z_t^{(e)}\right) = p_t^{(e)}, \\
\mathbb{P}\!\left(Y_t^{(e)} = 0 \mid Z_t^{(e)}\right) = 1 - p_t^{(e)}.
\end{gathered}
\label{eq:bernoulli}
\end{equation}
where $p_t^{(e)} \in [0,1]$ is the failure probability to be estimated.

The remaining task is to choose a parametric form for estimating $p_t^{(e)}$. Equipment failures are rare, which creates an imbalance between failure and non-failure observations and makes parameter estimation difficult. Furthermore, changes in equipment behavior are generally not isolated instantaneous events. When the true failure probability becomes high, it may remain elevated over several successive observation times. Similarly, under stable operating conditions, a low probability at time $t$ is usually consistent with low probability at recent previous times. This temporal persistence of failure probability means that successive predictions should not be treated as independent. A model that ignores this temporal structure may produce erratic probability estimates, alternating abruptly between high and low values without reflecting the gradual evolution of degradation.

The aim of this modeling step is therefore to construct a predictive model that satisfies three requirements. First, the output must lie in $[0,1]$, since it is a probability. Second, the model must remain robust when failure observations are rare. Third, it must account for the temporal persistence of failure probability.

Several learning-based methods could be used for this prediction task, including support vector machines, artificial neural networks, deep neural networks, and autoencoder-based models \cite{Datadriven2019IEEE,syed2025timeseriesreview}. Logistic regression is chosen among these alternatives for three reasons. First, it achieves competitive predictive accuracy while using few parameters compared to deep neural networks \cite{Datadriven2019IEEE}, making it a lightweight method suitable for rare-events. Second, it easily integrates past observations of different statistical types, including real-valued and binary variables, through the latent vector $Z_t^{(e)}$. Third, a prediction model based on logistic regression is reproducible and interpretable, which is essential for maintenance decision support.

Given the latent vector $Z_t^{(e)} \in \mathbb{R}^{d_z}$, the failure probability is first estimated using a basic logistic 
\begin{equation}
\widehat{p}_t^{(e)} = \sigma\!\left(b + \beta^\top Z_t^{(e)}\right),
\label{eq:logistic_model}
\end{equation}
where $\sigma(\cdot)$ is the logistic sigmoid function defined in Eq.~\eqref{eq:auxiliary_logistic_head}, $b \in \mathbb{R}$ is a scalar intercept, $\beta \in \mathbb{R}^{d_z}$ is the coefficient vector, and the parameter vector is $\theta^{\mathrm{log}} = (b, \beta^\top)^\top \in \mathbb{R}^{d_z+1}$.

However, this basic logistic model treats each time step as independent and does not exploit the temporal persistence of failure probability. The autologistic extension of logistic regression introduces dependence between neighboring binary response variables and has been used successfully to model dependent binary output, notably in epidemiological risk modeling \cite{betterautologistic,autologisticusecase1}. This motivates its adaptation to the present case, where the temporal persistence of failure probability creates dependence between successive predictions.

To obtain the autologistic model, the linear predictor of the basic logistic model in Eq.~\eqref{eq:logistic_model} is augmented by a weighted sum of previously predicted probabilities $\widehat{p}_{t-k}^{(e)}$, which are available at prediction time. Let $r \geq 0$ be the number of past predicted probabilities used by the model. We define the vector of autologistic parameters $a = (a_1, \ldots, a_r)^\top \in \mathbb{R}^r$, where $a_k \in \mathbb{R}$ is the coefficient associated with the predicted probability at lag $k$. The complete parameter vector is then $\theta = (b, \beta^\top, a^\top)^\top \in \mathbb{R}^{d_z+r+1}$.

The autologistic model is given by
\begin{equation}
\widehat{p}_t^{(e)} = \sigma\!\left(b + \beta^\top Z_t^{(e)} + \sum_{k=1}^{r} a_k\,\widehat{p}_{t-k}^{(e)}\right),
\label{eq:autologistic_model}
\end{equation}
When $r=0$, Eq.~\eqref{eq:autologistic_model} reduces to the basic logistic model in Eq.~\eqref{eq:logistic_model}. When $r>0$, the past predicted probabilities act as temporal covariates that allow recent probability estimates to influence the current prediction.

The autologistic term introduces temporal consistency by using recent predicted probabilities in the current prediction. It reduces abrupt fluctuations in the predicted failure probability and makes the prediction sequence more consistent with the gradual evolution of degradation.

\subsection{Parameter estimation}
\label{subsec:estimation}

For a given equipment $e$, parameter estimation is performed from the labeled latent observations $\bigl(Z_t^{(e)},Y_t^{(e)}\bigr)$. For a fixed parameter vector $\theta$, the model in Eq.~\eqref{eq:autologistic_model} produces a predicted failure probability $\widehat{p}_t^{(e)} \in [0,1]$ at each observation time $t=1,\ldots,N_e$. Using the Bernoulli model in Eq.~\eqref{eq:bernoulli}, the log-likelihood is written as

\begin{equation}
\ell(\theta)
= \sum_{t=1}^{N_e}
\Bigl[
Y_t^{(e)} \log \widehat{p}_t^{(e)}
+ \bigl(1 - Y_t^{(e)}\bigr) \log\bigl(1 - \widehat{p}_t^{(e)}\bigr)
\Bigr].
\label{eq:log_likelihood_single}
\end{equation}
In Eq.~\eqref{eq:log_likelihood_single} all observations have equal weights. In the present setting, failures are rare events, so the sum is dominated by non-failure observations, which may lead the model to underestimate the failure probability. To reduce this effect, we introduce class weights $\omega_1 > 0$ and $\omega_0 > 0$ assigned to failure and non-failure observations, respectively. Let $\Omega_1^{(e)}$ and $\Omega_0^{(e)}$ denote the sets of
time indices associated with failure and non-failure observations,
respectively,
\begin{equation}
\begin{aligned}
\Omega_1^{(e)}
&=
\left\{t \in \{1,\ldots,N_e\} \mid Y_t^{(e)}=1\right\},
\\
\Omega_0^{(e)}
&=
\left\{t \in \{1,\ldots,N_e\} \mid Y_t^{(e)}=0\right\}.
\end{aligned}
\label{eq:class_index_sets}
\end{equation}
The weighted log-likelihood is then
\begin{equation}
\ell_w(\theta)
=
\omega_1
\sum_{t\in\Omega_1^{(e)}}
\log \widehat{p}_t^{(e)}
+
\omega_0
\sum_{t\in\Omega_0^{(e)}}
\log\!\left(1-\widehat{p}_t^{(e)}\right).
\label{eq:weighted_log_likelihood_single}
\end{equation}

Class weighting reduces the effect of imbalance, but does not guarantee more precise estimators when the number of failure observations is small. We therefore introduce a Gaussian prior distribution on the model parameters. The prior is placed on the parameter vectors $\beta$ and $a$, while the intercept $b$ is left unpenalized so that the baseline failure probability can be estimated freely. Specifying this prior requires choosing a covariance structure. To keep the number of hyperparameters small, we assume independent components. A full covariance matrix would allow correlations between parameters, but would require estimating many additional hyperparameters. This is not appropriate in the present rare-event context. Formally, the Gaussian prior is
\begin{equation}
(\beta^\top, a^\top)^\top
\sim
\mathcal{N}\left(
0,
\sigma^2 I_{d_z+r}
\right),
\label{eq:prior}
\end{equation}
where $I_{d_z+r}$ is the identity matrix of size $(d_z+r) \times (d_z+r)$, and $\sigma^2 > 0$ is a hyperparameter controlling the strength of regularization.

Combining the weighted log-likelihood with the Gaussian prior gives the log-posterior, up to an additive constant. This posterior can be used in different ways, for example by sampling it with Markov chain Monte Carlo methods or by approximating it with variational inference \cite{geman1984stochastic,blei2017variational}. In this paper, we use maximum a posteriori (MAP) estimation, because the objective is to obtain a single set of prediction parameters equipment. The MAP estimator is given by
\begin{equation}
\widehat{\theta}^{\,\mathrm{MAP}}
= \operatorname*{arg\,max}_{\theta}
\left\{
\ell_w(\theta)
- \frac{1}{2\sigma^2}
\bigl(\|\beta\|_2^2 + \|a\|_2^2\bigr)
\right\}.
\label{eq:map_single}
\end{equation}
The resulting estimator is devoted to the prediction of failures for a single equipment.

\section{Common model and target-specific adaptation}
\label{sec:global_local}

The MAP estimator defined in Eq.~\eqref{eq:map_single} is formulated for a single equipment and uses only the data available for that equipment. This can be limiting when few failure observations are available on this equipment. We therefore estimate a common model using the training equipment from the same family, and then specialize this model for each target equipment. Let $\mathcal{E}$ denote the set of training equipment. The subscript $g$ denotes this common model, estimated from $\mathcal{E}$. The single-equipment weighted log-likelihood of Eq.~\eqref{eq:weighted_log_likelihood_single} is extended to all equipment in $\mathcal{E}$ by summing over this set:

\begin{equation}
\ell_w(\theta_g)
= \omega_1 \sum_{e \in \mathcal{E}}
            \sum_{t \in \Omega_1^{(e)}} \log \widehat{p}_{g,t}^{(e)}
+ \omega_0 \sum_{e \in \mathcal{E}}
            \sum_{t \in \Omega_0^{(e)}}
            \log\bigl(1 - \widehat{p}_{g,t}^{(e)}\bigr),
\label{eq:weighted_log_likelihood_global}
\end{equation}
where $\widehat{p}_{g,t}^{(e)}$ is obtained from Eq.~\eqref{eq:autologistic_model} with the common parameter vector $\theta_g = (b_g,\, \beta_g^\top,\, a_g^\top)^\top$. With the same Gaussian prior as in Eq.~\eqref{eq:prior}, the common MAP estimate is
\begin{equation}
\widehat{\theta}_g^{\,\mathrm{MAP}}
= \operatorname*{arg\,max}_{\theta_g}
\left\{
\ell_w(\theta_g)
- \frac{1}{2\sigma_g^2}
\bigl(\|\beta_g\|_2^2 + \|a_g\|_2^2\bigr)
\right\}.
\label{eq:common_map}
\end{equation}

The optimization problem in Eq.~\eqref{eq:common_map} is solved using the L-BFGS-B algorithm, a quasi-Newton method well-suited for smooth, unconstrained or bound-constrained optimization \cite{saputro2017lbfgs}. The algorithm is iterative; the parameters are initialized at zero, except for the intercept $b_g$ which is initialized using the empirical log-odds of the common failure rate.

When a new equipment is encountered, the common parameters provide a reasonable starting point and reduce the risk of overfitting to limited target data. The common parameters $\widehat{\beta}_g$ and $\widehat{a}_g$ are therefore used as prior means in the target-specific model. This is because the common model captures failure-related patterns shared across the equipment family.

For the specialization of the model to a target equipment $e^*$ unseen during common model training, the autologistic prediction parameters $\theta_{e^*} = (b_{e^*},\, \beta_{e^*}^\top,\, a_{e^*}^\top)^\top$ are re-estimated. The common model parameters serve as an informative prior:
\begin{equation}
(\beta_{e^*}^\top, a_{e^*}^\top)^\top
\;\Big|\;
\widehat{\theta}_g^{\,\mathrm{MAP}}
\sim
\mathcal{N}\!\left(
(\widehat{\beta}_g^\top, \widehat{a}_g^\top)^\top,\,
\sigma_{e^*}^2\, I_{d_z+r}
\right).
\label{eq:target_prior}
\end{equation}
The intercept $b_{e^*}$ is left unpenalized so that the target equipment can freely adjust its baseline failure probability. Let $\widehat{p}_{\mathrm{ad},t}^{(e^*)}$ denote the predicted failure probability for target equipment $e^*$ after adaptation, obtained with the target-specific parameters $\theta_{e^*}$. The adapted MAP estimate is

\begin{equation}
\begin{aligned}
\widehat{\theta}_{e^*}^{\,\mathrm{MAP}}
&=
\operatorname*{arg\,max}_{\theta_{e^*}}
\Biggl\{
\omega_1 \sum_{t \in \Omega_1^{(e^*)}}
\log \widehat{p}_{\mathrm{ad},t}^{(e^*)}
+
\omega_0 \sum_{t \in \Omega_0^{(e^*)}}
\log\!\left(1-\widehat{p}_{\mathrm{ad},t}^{(e^*)}\right)
\\[-2pt]
&\quad -
\frac{1}{2\sigma_{e^*}^2}
\left(
\left\|\beta_{e^*}-\widehat{\beta}_g\right\|_2^2
+
\left\|a_{e^*}-\widehat{a}_g\right\|_2^2
\right)
\Biggr\}.
\end{aligned}
\label{eq:target_map}
\end{equation}

The optimization problem in Eq.~\eqref{eq:target_map} is also solved using the L-BFGS-B algorithm \cite{saputro2017lbfgs}. The parameters are initialized at the common MAP estimate $\widehat{\theta}_g^{\,\mathrm{MAP}}$, which provides a warm start close to the optimum. This initialization is particularly beneficial when the target equipment contains few failure observations. The common parameters already encode the shared degradation structure, and the target data need only adjust the decision boundary to the specific characteristics of the new equipment.

Equation~\eqref{eq:target_map} balances target-specific adaptation and regularization toward the common model. The target parameters can move away from the common parameters only when the target observations provide enough evidence for such a deviation.
\section{Dataset}
\label{sec:simulator}

Evaluating a transferable failure-prediction model requires data that satisfy several conditions simultaneously: multiple refrigerators from the same equipment family, heterogeneous sensor configurations, recorded failure occurrence times with pre-failure degradation periods, and sufficient variability in operating conditions to evaluate generalization from training refrigerators to target refrigerators. Publicly available datasets for household equipment rarely satisfy all these conditions simultaneously \cite{nunes2023manufacturing}. We therefore generate a controlled synthetic dataset based on refrigerators. Refrigerators are used as an example of family of equipment because they operate continuously, produce time-dependent sensor-output data, and share the same basic function and physical principles while still exhibiting variability in sensor configuration, ambient conditions, usage, and degradation mechanisms. This makes them suitable for evaluating whether a model learned from several refrigerators can be applied to target refrigerators that were not used during training. The dataset consists of 27 simulated refrigerators equipped with 15 to 20 sensors, each recorded at one-minute resolution. The sensors measure quantities related to thermal behavior, electrical consumption, mechanical activity, and component state, including temperature, humidity, pressure, electrical current, power and energy consumption, vibration, fan speed, duty cycle, acoustic activity, frost thickness, door activity, compressor state, and defrost state. Not all refrigerators share the same sensor list, and the measured quantities differ in unit, range, and scale; for example, temperature is measured in $^{\circ}$C, pressure in bar, vibration in $g$, acoustic activity in dB, and fan speed in RPM. For each refrigerator $e$ and observation time $t$, the input vector is formed as $\Phi_t^{(e)} = [\widetilde{X}_t^{(e)\top}, M_t^{(e)\top}, C_t^{(e)\top}]^\top$, with dimension $d_\Phi = 2*me + d_c = 2\times25+15 = 65$. The binary label $Y_t^{(e)}$ is derived from the recorded failure occurrence times $t_{e,j}^{\mathrm{fail}}$ according to Eq.~\eqref{eq:future_target}.

Each simulated failure evolves through three successive periods. During the normal period, the sensor outputs indicate fault-free operation. This is followed by a degradation period, during which one or more sensor outputs gradually change before the failure becomes observable. The failure period then begins at the failure occurrence time $t_{e,j}^{\mathrm{fail}}$. In this dataset, six failure types are simulated: compressor degradation, mechanical wear, condenser fouling, door-seal degradation, fan degradation, and defrost-heater degradation. These failure types do not represent all possible refrigerator failures; they are selected to generate diverse pre-failure patterns in thermal, electrical, mechanical, and state-related sensor-output time series. The data-generation procedure, based on physical laws describing refrigerator behavior, is described in \cite{benamirouche2026syntheticdataset}.

The dataset is divided into a training set $\mathcal{E}_{\mathrm{train}}$ of 17 refrigerators and a target set $\mathcal{E}_{\mathrm{tgt}}$ of 10 refrigerators. The training set is used to estimate the common model, including the supervised autoencoder, its encoder $f_{\mathrm{enc}}$ defined in Eq.~\eqref{eq:encoder}, and the common MAP parameters $\widehat{\theta}_g^{\,\mathrm{MAP}}$ obtained from Eq.~\eqref{eq:common_map}. The target set is excluded from this training phase and is reserved for adaptation and evaluation. All refrigerators are represented in the same common sensor space, but each refrigerator observes only a subset of this space. Therefore, some sensor outputs are available for a given refrigerator, while others are absent. The target refrigerators vary in operating context, ambient conditions, compressor characteristics, sensor availability, and number of failures (see Table~\ref{tab:target_equipment_characteristics}), providing a heterogeneous testbed for assessing common-model generalization and target-specific adaptation.

\begin{table}[!htbp]
\centering
\caption{Characteristics of the 10 target refrigerators. Ambient temperature is given in $^{\circ}\mathrm{C}$ and electrical current in amperes (A). The sensor column lists additional sensors available for each refrigerator beyond the common base set of sensors.}
\label{tab:target_equipment_characteristics}

\footnotesize
\setlength{\tabcolsep}{4pt}
\renewcommand{\arraystretch}{1.15}

\begin{tabular}{L{0.8cm} L{1.9cm} L{2.0cm} L{1.8cm} L{3.6cm} L{1.4cm}}
\toprule
\textbf{Ref.} &
\textbf{Usage} &
\textbf{Ambient temp.} &
\textbf{Current} &
\textbf{Additional sensors} &
\textbf{Failures} \\
\midrule
T01 & Residential      & $22 \pm 3\,^{\circ}\mathrm{C}$   & $2.0\,\mathrm{A}$        & noise, vibration & 6 \\
T02 & Commercial       & $25 \pm 4\,^{\circ}\mathrm{C}$   & $2.8\,\mathrm{A}$        & suction pressure, discharge pressure & 4 \\
T03 & Residential      & $20 \pm 2\,^{\circ}\mathrm{C}$   & $1.6\,\mathrm{A}$        & vibration, humidity, frost thickness & 5 \\
T04 & Restaurant       & $26 \pm 4\,^{\circ}\mathrm{C}$   & $3.2\,\mathrm{A}$        & vibration, suction pressure, discharge pressure, door-opening count & 6 \\
T05 & Office           & $20 \pm 2.5\,^{\circ}\mathrm{C}$ & $1.4\,\mathrm{A}$        & humidity & 4 \\
T06 & Industrial       & $12 \pm 5\,^{\circ}\mathrm{C}$   & $4.2\,\mathrm{A}$        & vibration, suction pressure, discharge pressure & 9 \\
T07 & Residential      & $21 \pm 3\,^{\circ}\mathrm{C}$   & $1.5$--$3.0\,\mathrm{A}$ & noise & 6 \\
T08 & Commercial       & $32 \pm 5\,^{\circ}\mathrm{C}$   & $3.8\,\mathrm{A}$        & vibration, humidity & 5 \\
T09 & Residential      & $22 \pm 3\,^{\circ}\mathrm{C}$   & $2.4\,\mathrm{A}$        & noise, vibration, door-opening count & 8 \\
T10 & Light commercial & $23 \pm 3.5\,^{\circ}\mathrm{C}$ & $2.2\,\mathrm{A}$        & noise, vibration, suction pressure, discharge pressure & 6 \\
\midrule
\multicolumn{5}{l}{Total} & 59 \\
\bottomrule
\end{tabular}
\end{table}
Performance measures on each target refrigerator in two successive stages:

\begin{enumerate}[(i)]
\item \textbf{Common model prediction.} The common model parameters $\widehat{\theta}_g^{\mathrm{MAP}}$ estimated using Eq.~\eqref{eq:common_map} are applied directly to $\mathcal{E}_{\mathrm{tgt}}$. 

\item \textbf{Target-specific model.} For a target refrigerator $e^*$, the parameters $\theta_{e^*} = (b_{e^*},\,\beta_{e^*}^\top,\,a_{e^*}^\top)^\top$ are estimated via Eq.~\eqref{eq:target_map}, using $\widehat{\theta}_g^{\mathrm{MAP}}$ as prior information, and the resulting model is evaluated on the corresponding target refrigerator.
\end{enumerate}

\section{Results and discussion}
\label{sec:results}

\subsection{Evaluation metrics}

The output of the model at each time $t$ is the predicted failure probability $\widehat{p}_t^{(e)}$ from Eq.~\eqref{eq:autologistic_model}. An alarm is active at time $t$ when $\widehat{p}_t^{(e)} \geq \tau$, where $\tau \in [0,1]$ is a fixed decision threshold for each equipment $e$.

The evaluation is performed at the event level, relative to each failure occurrence time $t_{e,j}^{\mathrm{fail}}$. Specifically, the $j$-th failure of equipment $e$ is counted as detected if there exists a time $t_{e,j}^{\mathrm{det}} < t_{e,j}^{\mathrm{fail}}$ such that the alarm remains continuously active from $t_{e,j}^{\mathrm{det}}$ until the failure occurrence,
\begin{equation}
\widehat{p}_t^{(e)} \geq \tau,
\qquad
\forall\, t \in
\bigl[t_{e,j}^{\mathrm{det}},\; t_{e,j}^{\mathrm{fail}}\bigr].
\label{eq:detection_rule}
\end{equation}
Conversely, if the predicted probability drops below $\tau$ at any time in this interval, the alarm is interrupted and the failure is counted as missed. 

For a detected failure, the lead time is defined as
\begin{equation}
L_{e,j} = t_{e,j}^{\mathrm{fail}} - t_{e,j}^{\mathrm{det}}.
\label{eq:lead_time}
\end{equation}
Thus, lead time measures the time between the first valid alarm and the failure occurrence. A longer lead time indicates that the detected failure is anticipated earlier, giving more time to plan maintenance. The mean lead time reported in the results is then the average of lead times over all target equipments.

In parallel, a false alert is defined as a time interval during which $\widehat{p}_t^{(e)} \geq \tau$ but no failure occurs within the subsequent horizon $]t,\, t+H]$. Each such interval is counted as one false alert, regardless of its duration.

Based on these definitions, the detection rate summarizes event-level performance as the fraction of failures that are detected
\begin{equation}
\mathrm{DR}
=
\frac{\text{number of detected failures}}
     {\text{total number of failures}}.
\label{eq:detection_rate}
\end{equation}

In addition to these event-level metrics, the area under the ROC curve (AUC--ROC) is reported as a complementary measure. Indeed, it evaluates how well the predicted probabilities separate observations followed by a failure from observations not followed by a failure, by varying the decision threshold over all possible values. Consequently, an AUC--ROC of 1 indicates perfect separation, an AUC--ROC of 0.5 indicates no discriminative ability, and an AUC--ROC below 0.5 indicates that the predictions are worse than random.

Similarly, average precision (AP) is used to select the latent-space dimension $d_z$. For each candidate value, the supervised autoencoder produces latent vectors $Z_t^{(e)}$, and AP summarizes the precision--recall curve over all thresholds. Specifically, precision measures the fraction of alarms that correspond to actual failures, so high precision means few false alerts, whereas recall measures the fraction of actual failures that are detected, so high recall means few missed failures. Therefore, AP balances these two objectives and ranges from 0 to 1. A higher AP thus indicates a better trade-off between limiting false alerts and detecting failures.

Finally, to improve the stability of predicted probabilities, an optional post-processing step is applied. The autologistic model in Eq.~\eqref{eq:autologistic_model} already introduces temporal regularity through the $r$ most recent predicted probabilities. However, residual short-term fluctuations in $\widehat{p}_t^{(e)}$ can briefly drop the alarm below $\tau$ within a pre-failure interval, causing a missed failure under Eq.~\eqref{eq:detection_rule}, or briefly raise it above $\tau$ during normal operation, causing a spurious alert. To attenuate both effects, a moving average over a window of $W$ time steps is therefore applied after prediction,
\begin{equation}
\widetilde{p}_t^{(e)} = \frac{1}{W} \sum_{k=0}^{W-1} \widehat{p}_{t-k}^{(e)}.
\label{eq:smoothed_prob}
\end{equation}
This window is wider than the autologistic memory of $r$ steps, hence it smooths fluctuations that the autologistic term does not suppress. When smoothing is applied, $\widetilde{p}_t^{(e)}$ then replaces $\widehat{p}_t^{(e)}$ in the alarm rule and in Eq.~\eqref{eq:detection_rule}. The value of $W$ is set between 0 and 120 depending on the equipment.

\subsection{Performance assesement on target refrigerators}

In the following table, we present the performance of the common model and the target-specific model, and then we compare these two prediction models across the 10 target refrigerators one by one.

\begin{table}[!htbp]
\centering
\caption{Performance comparison on the 10 target refrigerators.}
\label{tab:performance_comparison}
\begin{tabular}{lcc}
\toprule
Metric & Common model & Target-specific model \\
\midrule
Failures & 59 & 59 \\
Detected failures & 36 & 54 \\
Missed failures & 23 & 5 \\
Detection rate (\%) & 61.0 & 91.5 \\
False alerts & 24 & 13 \\
Mean lead time (h) & 132.8 & 101.2 \\
\bottomrule
\end{tabular}
\end{table}

With the common model, 36 out of 59 failures are detected, which corresponds to a detection rate of 61\%. This shows that the model learned from the training refrigerators can already detect a substantial part of the failures on target refrigerators whose data were not used to estimate the common parameters. However, 23 failures are still missed, which shows that the common model alone is not sufficient for reliable prediction on all target refrigerators. With the target-specific model, the number of detected failures increases from 36 to 54, the number of missed failures decreases from 23 to 5, and the detection rate increases from 61\% to 91.5\%. The number of false alerts also decreases from 24 to 13. Thus, the target-specific model improves failure detection while also reducing the total number of false alerts. The mean lead time decreases from 132.8 hours to 101.2 hours. This means that alarms start closer to the actual failure time, while still providing more than four days of advance warning on average. Overall, these results show that the common model provides a useful initial prediction model, while the target-specific model substantially improves failure detection on the target refrigerators.

\begin{figure}
\centering
\includegraphics[width=0.8\linewidth]{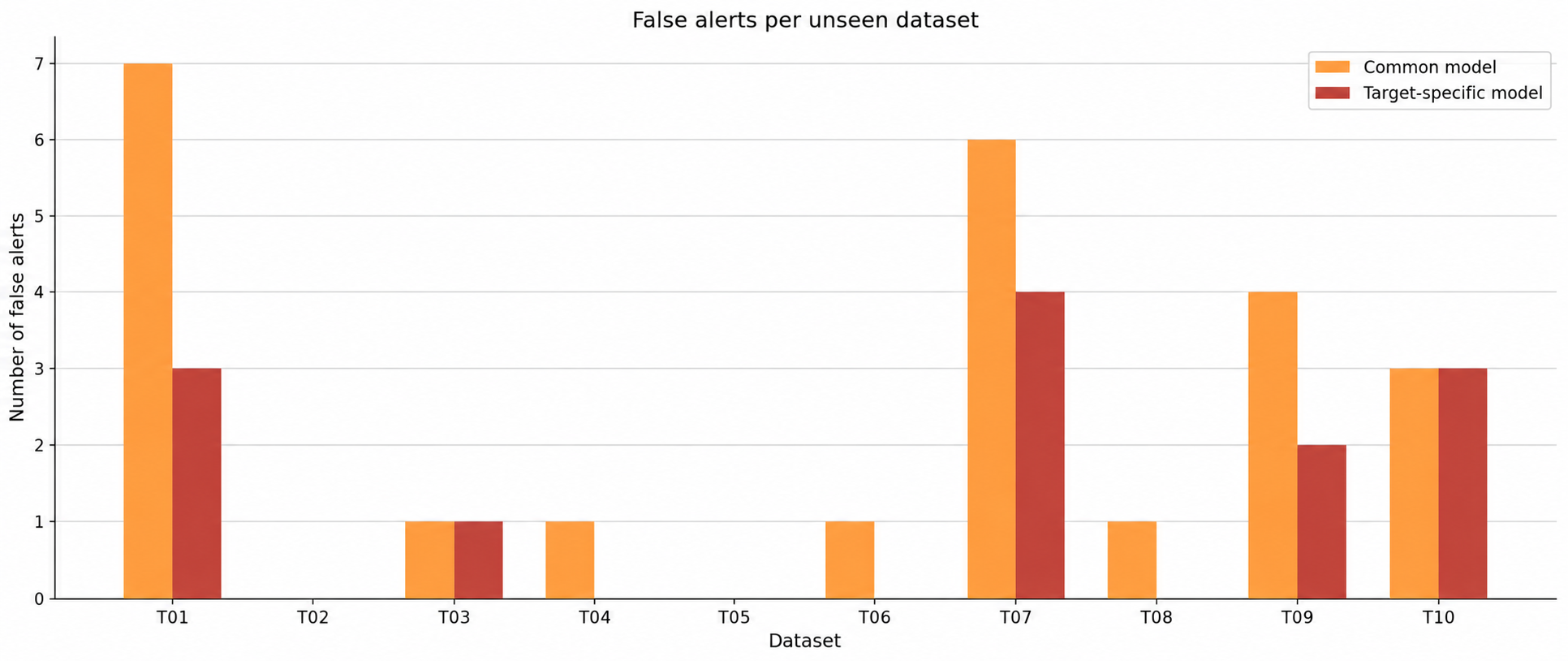}
\caption{False alerts on the 10 target refrigerators for the common model
and the target-specific model.}
\label{fig:Figure_1}
\end{figure}

This results in Table~\ref{tab:performance_comparison} show that the target-specific model reduces the total number of false alerts. To verify whether this reduction is concentrated on a few refrigerators or observed across several targets, Fig.~\ref{fig:Figure_1} reports the number of false alerts for each target refrigerator.

\subsection{Performance by target equipment}

Table~\ref{tab:per_equipment} gives the same comparison separately for each target refrigerator. The purpose is to verify whether the improvement of the target-specific model is observed for all target refrigerators or only for a small number of cases.

\begin{table}[t]
\centering
\caption{Performance comparison for each target refrigerator. 
Detected: detected failures; Missed: missed failures; 
False alerts: number of false alert intervals; 
$\tau_c$ and $\tau_t$: thresholds chosen per equipment to maximize detection
while minimizing false alerts; $\Delta$DR: difference in detection rate 
(percentage points).}
\label{tab:per_equipment}
\resizebox{\linewidth}{!}{%
\begin{tabular}{lc cccc c cccc c}
\toprule
& & \multicolumn{4}{c}{Common model} & &
\multicolumn{4}{c}{Target-specific model} & \\
\cmidrule(lr){3-6} \cmidrule(lr){8-11}
Target & Failures
& $\tau_c$ & Detected & Missed & False alerts
&
& $\tau_t$ & Detected & Missed & False alerts
& $\Delta$DR (\%) \\
\midrule
T01 & 6 & 0.50 & 2 & 4 & 7 & & 0.631 & 5 & 1 & 3 & $+50.0$ \\
T02 & 4 & 0.50 & 2 & 2 & 0 & & 0.500 & 3 & 1 & 0 & $+25.0$ \\
T03 & 5 & 0.50 & 3 & 2 & 1 & & 0.568 & 4 & 1 & 1 & $+20.0$ \\
T04 & 6 & 0.50 & 5 & 1 & 1 & & 0.500 & 6 & 0 & 0 & $+16.7$ \\
T05 & 4 & 0.65 & 2 & 2 & 0 & & 0.500 & 3 & 1 & 0 & $+25.0$ \\
T06 & 9 & 0.72 & 4 & 5 & 1 & & 0.500 & 9 & 0 & 0 & $+55.6$ \\
T07 & 6 & 0.50 & 5 & 1 & 6 & & 0.600 & 6 & 0 & 4 & $+16.7$ \\
T08 & 5 & 0.54 & 4 & 1 & 1 & & 0.500 & 5 & 0 & 0 & $+20.0$ \\
T09 & 8 & 0.70 & 6 & 2 & 4 & & 0.730 & 8 & 0 & 2 & $+25.0$ \\
T10 & 6 & 0.50 & 3 & 3 & 3 & & 0.520 & 5 & 1 & 3 & $+33.3$ \\
\midrule
Total & 59 & -- & 36 & 23 & 24 & & -- & 54 & 5 & 13 & $+30.5$ \\
\bottomrule
\end{tabular}}
\end{table}

The common model detects failures on all target refrigerators, but its performance varies. The detection rate ranges from 33\% on T01 (2 out of 6) and 44\% on T06 (4 out of 9) to 83\% on T04 and T07 (5 out of 6). This shows that the common model generalizes to target refrigerators, yet with uneven reliability. The target-specific model improves detection on every unit. The largest gain is on T06, where the detection rate rises from 44\% to 100\%. Complete detection is also reached on T04, T07, T08, and T09, with rates increasing from 83\% to 100\% on T04 and T07, from 80\% to 100\% on T08, and from 75\% to 100\% on T09. Smaller but consistent gains are observed on the remaining units, for example from 33\% to 83\% on T01 and from 50\% to 75\% on T02 and T05. Overall, the number of missed failures drops from 23 to 5.

The number of false alerts is also reduced after target-specific adaptation. It decreases on T01, T04, T06, T07, T08, and T09, and remains unchanged on T02, T03, T05, and T10. No target refrigerator shows an increase in false alerts. This confirms that the model does not detect more failures simply by raising more alarms. The target-specific model detects more failures while producing fewer false alerts overall.

\subsection{AUC--ROC comparison}
\label{subsec:Figure_2_3}

\begin{figure}
\centering
\begin{subfigure}{0.48\linewidth}
\centering
\includegraphics[width=\linewidth]{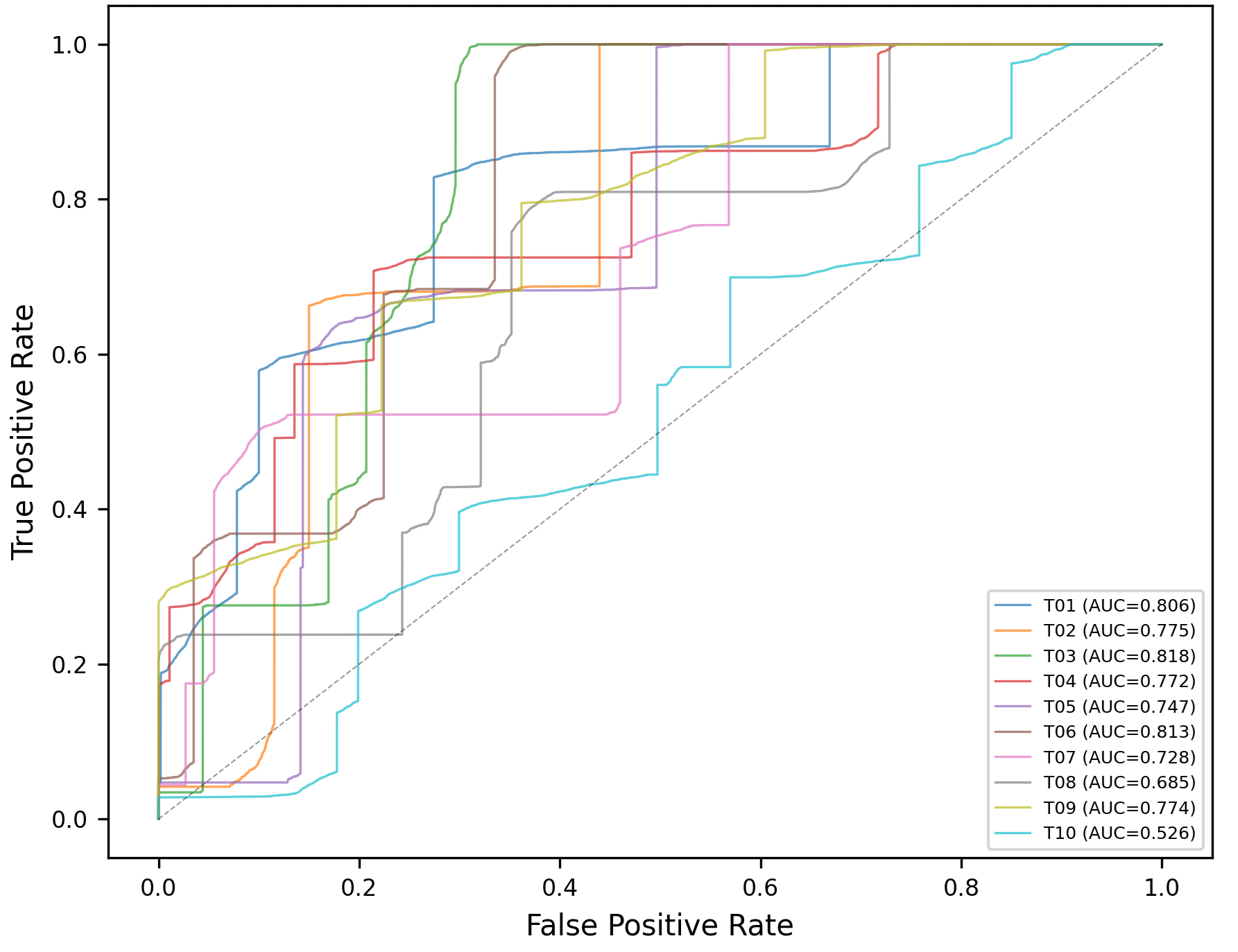}
\caption{Common model.}
\label{fig:Figure_2}
\end{subfigure}
\hfill
\begin{subfigure}{0.48\linewidth}
\centering
\includegraphics[width=\linewidth]{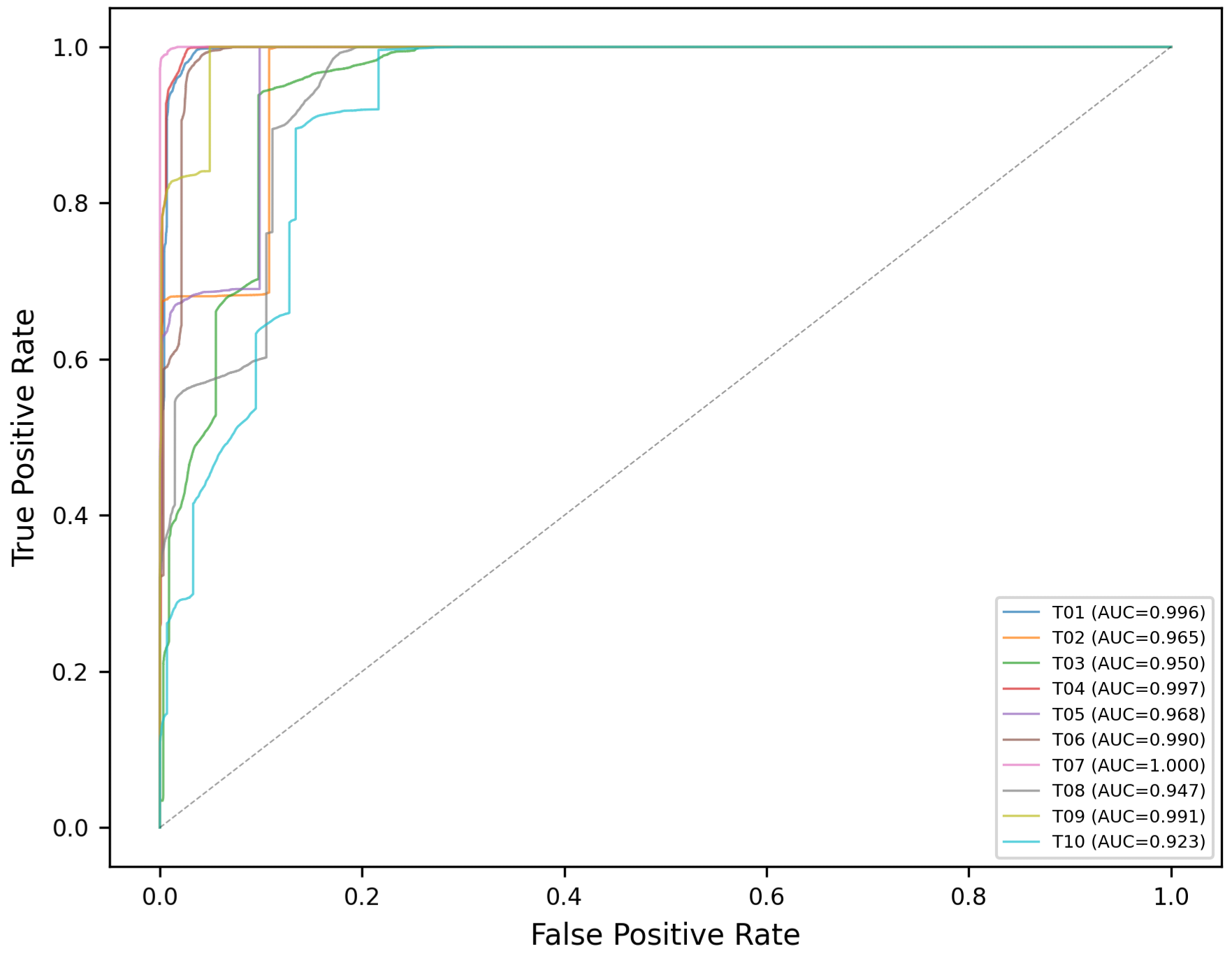}
\caption{Target-specific model.}
\label{fig:Figure_3}
\end{subfigure}
\caption{ROC curves on the 10 target refrigerators. The values in the
legends are AUC--ROC scores. AUC--ROC is reported as a complementary
observation-level measure and must be interpreted together with the
event-level metrics.}
\label{fig:Figure_2_3}
\end{figure}

Figure~\ref{fig:Figure_2_3} compares the ROC curves obtained on the 10 target refrigerators, with Figure~\ref{fig:Figure_2_3}a showing the common model applied without adaptation and Figure~\ref{fig:Figure_2_3}b showing the target-specific model after adaptation using Eq.~\eqref{eq:target_map}. The values in the legends are the AUC--ROC scores. With the common model, these scores range from 0.526 to 0.818, indicating moderate and variable separation between observations followed by a failure and those not followed by a failure. Some targets such as T01, T03, and T06 score above 0.80, while T10 remains near random performance at 0.526. After target-specific adaptation, all ROC curves shift toward the upper-left corner and the AUC--ROC scores rise to a range of 0.923 to 1.000, with the largest gain on T10 increasing from 0.526 to 0.923. This confirms that target-specific adaptation improves the observation-level separation between observations followed by a failure and observations not followed by a failure. However, AUC--ROC remains a complementary measure. It summarizes separation quality across all thresholds but does not indicate how many failures are detected, how many are missed, how many false alerts are produced, or how early valid alarms are raised. The AUC--ROC analysis therefore supports the event-level results without replacing them.

\subsection{Selection of the latent dimension}

\begin{figure}
\centering
\includegraphics[width=0.8\linewidth]{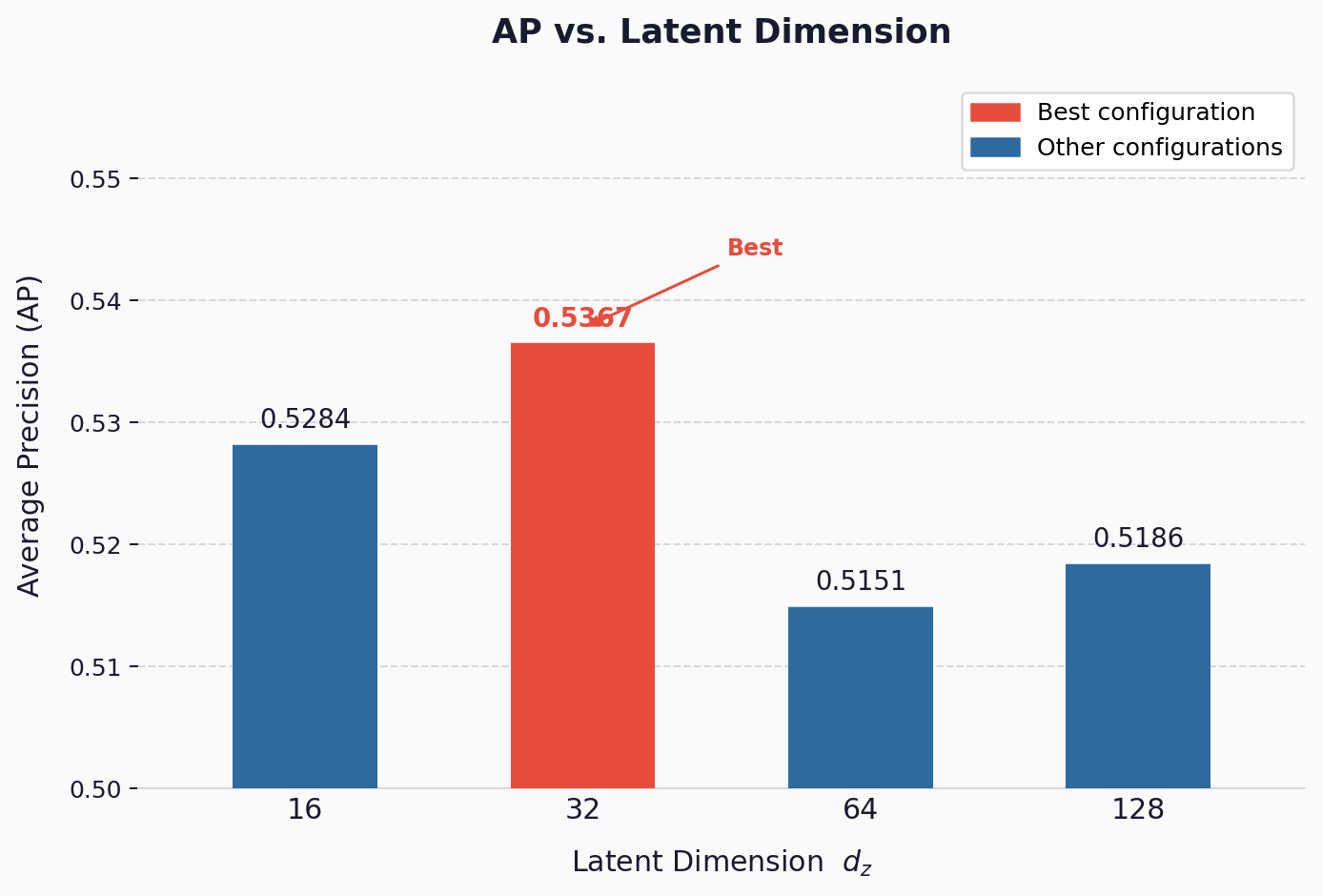}
\caption{Average precision (AP) obtained for different latent dimensions
$d_z$. The best validation result is obtained with $d_z=32$.}
\label{fig:Figure_4}
\end{figure}

Figure~\ref{fig:Figure_4} displays four dimensions for the latent vector $Z_t^{(e)}$, with $d_z \in \{16, 32, 64, 128\}$. The selection is based on validation average precision (AP), which balances failure detection and false alerts. The best result is obtained with $d_z = 32$ (AP = 0.5367). The other dimensions give lower AP values: 0.5284 for $d_z = 16$, 0.5151 for $d_z = 64$, and 0.5186 for $d_z = 128$. Therefore, $d_z = 32$ is retained for all subsequent experiments.

\subsection{Discussion}

The experimental results support three main observations. First, the common model generalizes without using target data. It detects 36 out of 59 failures, corresponding to a detection rate of 61\%. This shows that the encoder and the common autologistic parameters capture transferable failure-related patterns from the training refrigerators. However, 39\% of failures are still missed, confirming that the common model alone is not sufficient for reliable prediction on all targets.

Second, target-specific adaptation substantially improves detection. The number of detected failures increases from 36 to 54, and missed failures decrease from 23 to 5. The detection rate therefore rises from 61\% to 91.5\%. This improvement is observed on every target refrigerator. Since only the final prediction layer is adapted, the latent vector learned from the common model provides a useful basis, but the prediction parameters must be adjusted to each target.

Third, the detection gain does not come at the cost of more false alerts. On the contrary, false alerts decrease from 24 to 13, a reduction of 46\%. The mean lead time decreases from 132.8 hours to 101.2 hours, meaning alarms start closer to the failure time while still providing more than four days of advance warning on average. The remaining limitations are the 5 missed failures (8.5\%) and the 13 false alerts. Future work should investigate improved alarm-generation rules to further reduce both.

\section{Conclusion}
\label{sec:conclusion}

This paper proposed a common-to-target probabilistic model for predicting rare failures in heterogeneous equipment monitored by multivariate sensor-output time series, where the equipment behavior is treated as a random process whose failure probability depends on recent sensor history and operating context. The solution represents heterogeneous data using a common sensor space, availability masks, and a context vector, then learns a latent vector of recent equipment data through a supervised encoder. This vector feeds an autologistic model that estimates failure probability while accounting for temporal persistence, and the common model learned from training equipment is specialized to each target equipment through adaptation. Experiments on synthetic refrigerator data generated from physical laws describing thermal and mechanical behavior show that the common model detects 61\% of failures on target equipment without using them during training, while target-specific adaptation raises this to 91.5\% and reduces false alerts by nearly half. These results confirm that the model transfers failure-related patterns across heterogeneous equipment and improves prediction. Future work will focus on validation with real-world data, non-stationary operating conditions, and improved alarm-generation rules to further reduce missed failures and false alerts in rare-failures.

\section*{Data availability statement}

The synthetic dataset and the detailed data-generation procedure are not publicly available at the time of submission because they form the subject of a separate data article that is currently under preparation. The dataset and the corresponding generation documentation will be made publicly available in an appropriate repository upon publication of the data article.

\section*{CRediT authorship contribution statement}

\textbf{Islam Benamirouche:} Conceptualization, Methodology, Software, Formal analysis,
Writing -- original draft.

\textbf{Djemel Ziou:} Supervision, Conceptualization, Validation, Project administration,
Writing -- review and editing.

\textbf{Feriel Fass:} Validation, Writing -- review and editing.

\bibliographystyle{elsarticle-num}
\bibliography{article-refs}

\end{document}